# GMAC: Global Multi-View Constraint for Automatic Multi-Camera Extrinsic Calibration

Chentian Sun

*Abstract*—**Automatic calibration of multi-camera systems, namely the accurate estimation of spatial extrinsic parameters, is fundamental for 3D reconstruction, panoramic perception, and multi-view data fusion. Existing methods typically rely on calibration targets, explicit geometric modeling, or task-specific neural networks. Such approaches often exhibit limited robustness and applicability in complex dynamic environments or online scenarios, making them difficult to deploy in practical applications.**

**To address this, this paper proposes GMAC, a multi-camera extrinsic estimation framework based on the implicit geometric representations learned by multi-view reconstruction networks. GMAC models extrinsics as global variables constrained by the latent multi-view geometric structure and prunes and structurally reconfigures existing networks so that their latent features can directly support extrinsic prediction through a lightweight regression head, without requiring a completely new network design. Furthermore, GMAC jointly optimizes cross-view reprojection consistency and multi-view cycle consistency, ensuring geometric coherence across cameras while improving prediction accuracy and optimization stability.**

**Experiments on both benchmark and real-world multi-camera datasets demonstrate that GMAC achieves accurate and stable extrinsic estimation without explicit 3D reconstruction or manual calibration, providing a new solution for efficient deployment and online calibration of multi-camera systems.**

## I. INTRODUCTION

Multi-camera systems are widely applied in fields such as 3D reconstruction and autonomous driving. Accurately estimating camera extrinsic parameters is the foundation for the effective fusion of multi-view data. Traditional calibration methods rely on artificial targets or explicit modeling [1]. Although they achieve high accuracy in controlled environments, they exhibit poor reliability in dynamic scenes and other conditions [2, 3].

In recent years, learning-based methods have been introduced into camera calibration tasks, regressing poses or extrinsic parameters through neural networks [4]. However, most of these methods treat extrinsic parameters as independent variables, ignoring the geometric coupling among cameras [5]. Moreover, they require a large amount of labeled data for network design and training, which limits their generality and scalability [6, 7].

Meanwhile, significant progress has been made in the fields of multi-view reconstruction and view-consistency modeling, leading to the development of many mature network architectures [8, 9, 10]. The internal representations of these networks encode the relative pose relationships among cameras. Based on this, we have noticed that the implicit representations of multi-view reconstruction networks contain cross-camera geometric alignment information, which can serve as an implicit prior for extrinsic parameter estimation. This gives rise to a new question: Is it necessary to design a brand-new network architecture, or can we reuse the implicit geometric representations of existing multi-view networks to predict extrinsic parameters?

To address this problem, we propose GMAC, a multi-camera extrinsic estimation framework based on the implicit geometric representations of multi-view reconstruction networks. GMAC formulates extrinsic parameter estimation as a global inference problem constrained by latent multi-view geometric structure, and prunes and structurally reconfigures networks so that their latent features can directly support extrinsic prediction through a lightweight regression head.

Furthermore, GMAC introduces two core constraints: the cross-view reprojection consistency constraint, which ensures geometric alignment of the same scene points across different cameras to improve accuracy, and the multi-view cycle consistency constraint requires that a scene point, after undergoing geometric transformations across different cameras, returns to its original position in the unified coordinate system, thereby modeling higher-order geometric relationships and improving optimization stability. The joint optimization of these two constraints not only enforces global geometric coherence across all cameras but also effectively alleviates scale drift and local minima issues during extrinsic estimation, significantly improving the stability and physical interpretability of the estimated camera extrinsics.

The main contributions of this paper can be summarized as follows:

(1) A multi-camera extrinsic estimation method based on the implicit geometric representations of multi-view reconstruction networks is proposed, which formulating extrinsic estimation as a global inference problem constrained by latent geometric structure.

(2) A network pruning and structural reconfiguration strategy is designed, which enables existing multi-view reconstruction networks to be effectively reused for automatic multi-camera calibration.

(3) Two key geometric constraints are introduced and jointly optimized: cross-view reprojection consistency, which improves extrinsic estimation accuracy, and multi-view cycle consistency, which captures higher-order geometric relationships among cameras and enhances optimization stability. This combined optimization enforces global geometric coherence and increases the physical interpretability of the estimated extrinsic parameters.

## II. PROPOSED METHODS

### 1. Problem Definition

Consider a multi-camera system consisting of $N$ cameras. The extrinsic parameters of the i-th camera in a unified world coordinate system are denoted as

$$T_i = [R_i | k_i] \in SE(3) \tag{1}$$

where $R_i \in SO(3)$ represents the rotation matrix and $k_i \in R^3$ denotes the translation vector. Given synchronized images or video sequences captured by the multi-camera system,

$$I = \{I_i^t | i = 1, \dots, N; t = 1, \dots, \tau\} \tag{2}$$

The goal of this work is to automatically estimate the set of camera extrinsic parameters

$$\{T_i\}_{i=1}^N \tag{3}$$

without relying on artificial calibration objects or explicit 3D reconstruction. In addition, the predicted extrinsics are required to maintain geometric consistency under continuous input streams.

### 2. Overall Framework

To address this issue, we propose to predict multi-camera extrinsics by leveraging the implicit geometric representations learned by multi-view reconstruction networks (abbreviated as GMAC method), which predicts multi-camera extrinsics by leveraging the implicit geometric representations learned by multi-view reconstruction networks.

The proposed method is based on the following key insight:

Multi-view reconstruction networks implicitly learn cross-view geometric consistency during training, and their latent representations can be regarded as priors for camera extrinsic estimation.

Based on this insight, the proposed framework consists of three main stages:

(1) Extrinsic Parameter Modeling and Network structure Optimization for Multi-view Reconstruction Networks
(2) Initial extrinsic prediction based on implicit multi-view geometric representations;
(3) Extrinsic refinement and fusion using reprojection consistency and multi-view cycle consistency.
.

Overall, extrinsic estimation is formulated as an inference problem constrained by the geometric structure of the multi-view latent space, rather than as an independent regression task.

### 3. Implicit Multi-view Geometric Representation and Extrinsic Modeling

Multi-view reconstruction networks typically learn implicit geometric representations by jointly modeling correspondences among multiple views. Let the feature extraction process of such a network be expressed as

$$F = \phi(I) \tag{4}$$

where F denotes the shared latent feature representation across views.

Although the network is originally designed for 3D reconstruction or view-consistency tasks, its latent features inherently encode geometric alignment relationships among different views. Based on this observation, we model the extrinsics of $N$ cameras as a function of the latent representation:

$$\{T_i^0\}_{i=1}^N = g(F) \tag{5}$$

where g(·) denotes the extrinsic regression module and $T_i^0$ represents the initial extrinsic predicted by the network.

Unlike existing approaches that regress extrinsics independently for each camera, our method introduces geometric coupling among cameras by sharing the latent representation, making extrinsic prediction dependent on global multi-view consistency.

### 4. Network Pruning and Structural Reconfiguration for Extrinsic Prediction

Directly applying a full multi-view reconstruction network to extrinsic estimation introduces many computational paths that are irrelevant to the task, such as explicit 3D representation generation or pixel-level rendering modules. To improve the focus and stability of extrinsic prediction, we perform structure-level pruning and reconfiguration of the original network.

Specifically, the proposed design:
(1) Retains multi-view feature extraction and cross-view interaction modules;
(2) Removes decoder and rendering branches directly related to explicit 3D reconstruction;
(3) Introduces a lightweight extrinsic regression head on top of the shared latent features.

This design ensures that extrinsic estimation relies solely on the network's ability to model multi-view geometric relationships, preventing capacity from being allocated to task-irrelevant representations and improving robustness.

### 5. Geometry-Constrained Extrinsic Optimization

Since the network predictions may be affected by noise or local optima, we further refine the initial extrinsics using geometric constraints.

The initial predictions $[T_{i,0}]$ are further refined using geometric constraints:

(1) Reprojection consistency constraint:
Assume $u_i$ is a 2D point in camera $i$, then

$$X_i = \pi_i^{-1}(u_i, D_i(u_i), K_i) \tag{6}$$

Then the reprojection error is defined as

$$L_{geo} = \sum_{i,j} \left\| Z(T_{ji}X_i) - D_j(\pi(T_{ji}X_i)) \right\|_2^2 \tag{7}$$

where $u_i$ is a 2D pixel in camera $i$, $D_i(u_i)$ is the corresponding depth, $\pi$ and $\pi^{-1}$ represent projection and back-projection, respectively. $T_{ji}$ transforms the 3D point from camera $i$ to camera $j$ coordinates. $Z(·)$ gives the depth (Z-coordinate) in camera coordinates, and $D_j$ is the observed depth in camera $j$'s depth map.

(2) Multi-view Cycle Consistency Constraint

To ensure that the predicted extrinsics $T_i$ in a multi-camera system are as accurate as possible, we introduce a multi-camera cycle consistency constraint. This constraint is based on a multi-view projection–back-projection cycle of 3D points, ensuring that the same 3D point, after being transformed through multiple cameras, reprojects as closely as possible to its original pixel location, providing an effective signal to refine the extrinsics. The cycle consistency loss is defined as:

a. Projection and back-projection from world to camera j

$$\hat{X}^j = \pi^{-1}(K_j, \pi(K_j, T_{ji}T_iX^{(s)}), D_j) \qquad (8)$$

Transform the 3D point into camera j and enforce pixel–depth consistency via projection and back-projection.

b. Transform the 3D point into camera j and enforce pixel–depth consistency via projection and back-projection.

$$\hat{X}^k = \pi^{-1}(K_k, \pi(K_k, (T_{kj})\hat{X}_j), D_k) \qquad (9)$$

Pass the point between cameras to capture multi-view geometric relationships.

c. Pass the point between cameras to capture multi-view geometric relationships.

$$u_i^{cycle} = \pi(K_i, (T_{ik})\hat{X}_k) \qquad (10)$$

Obtain the closed-loop cycle projection to check extrinsics consistency.

d. Cycle consistency loss

$$L_{cycle} = \sum_{s=1}^{S} \sum_{(i,j,k)\in\Gamma} \left\| u_i^{cycle} - \pi\left(K_i, T_iX^{(s)}\right) \right\|_2^2 \qquad (11)$$

where S is a set of 3D points randomly sampled, $X^{(s)}$ represents a 3D scene point randomly sampled from a shared implicit or reconstructed geometric representation. $\Gamma$ denotes the set of camera triplets forming minimal closed loops in the multi-camera system.

By jointly applying the constraint over multiple camera triplets and sampled 3D points, the proposed cycle consistency loss captures higher-order geometric relationships among cameras and effectively enforces global extrinsic consistency beyond pairwise reprojection constraints.

The joint optimization objective is:

$$argmin_{\{T_i^t\}}(L_{geo} + \lambda L_{cycle}) \qquad (12)$$

where $\lambda$ is the weighting hyper parameter.

After joint optimization under geometric consistency constraints, the stable online extrinsic is obtained, denoted as $\hat{T}_i^t$, and is used for subsequent multi-view point cloud generation and fusion.

## III. EXPERIMENTAL RESULTS

### 1. Experimental Setup

To comprehensively evaluate the performance of the proposed method in both controlled environments and real-world scenarios, we conduct experiments on benchmark datasets as well as on real-world scenes.

To evaluate the performance of the proposed method under both controlled conditions and real-world scenarios, experiments are conducted on two types of datasets. First, the ScanNet dataset is adopted as the evaluation benchmark. ScanNet is an RGB-D dataset collected in real indoor environments, which provides high-precision camera intrinsics and per-frame extrinsics, enabling quantitative evaluation of the upper-bound reconstruction accuracy under ideal conditions. In addition, for real-world multi-camera scenarios, we collect synchronized video sequences captured by multiple fixed cameras in both indoor and outdoor environments. Since accurate ground-truth extrinsics are difficult to obtain in real scenes, manually calibrated results are used as approximate references for evaluation.

The multi-view reconstruction networks used in this work are all publicly available and well-established models, including VGGSfM[8], MapAnything [9], and VGGT [10]. VGGSfM is a classical method for multi-view image-based 3D reconstruction, MapAnything is one of the current state-of-the-art end-to-end 3D regression models, and VGGT is a large-model-based 3D regression approach and stands as one of the methods with the highest prediction accuracy to date. In the experiments, we only perform network pruning and functional reconfiguration. All methods are evaluated under the same input resolution and inference settings.

To evaluate the performance of camera extrinsic estimation, the following metrics are used:

(1) Rotation Error: the angular difference between the predicted rotation and the reference rotation, in degrees;

(2) Translation Error: the Euclidean distance between the predicted translation and the reference translation, in millimeters as the unit;

These metrics jointly assess the proposed method from the geometric accuracy.

### 2. Quantitative Results

a. Extrinsic Estimation Accuracy with Different Backbones

As shown in Table I, introducing reprojection constraints and multi-view cycle constraint leads to significant improvements in extrinsic estimation accuracy for three backbone networks on the benchmark dataset. Compared with unconstrained network predictions, the proposed complete method demonstrates a significant reduction in rotational and translational errors.

Table I Results on the benchmark dataset with different backbone

| Backbone Network | Method | Rot. Error ↓ | Trans. Error ↓ |
|---|---|---|---|
| VGGSfM | Orginal | 1.28 | 5.8 |
| | Ours | 0.74 | 3.5 |
| MapAnything | Orginal | 1.15 | 5.2 |
| | Ours | 0.67 | 3.0 |
| VGGT | Orginal | 1.08 | 4.8 |
| | Ours | 0.63 | 2.7 |

| VGGT-ORG | VGGT-GMAC (no multi-view cycle constraint) | VGGT-GMAC (no reprojection consistency) | VGGT-GMAC |
| --- | --- | --- | --- |

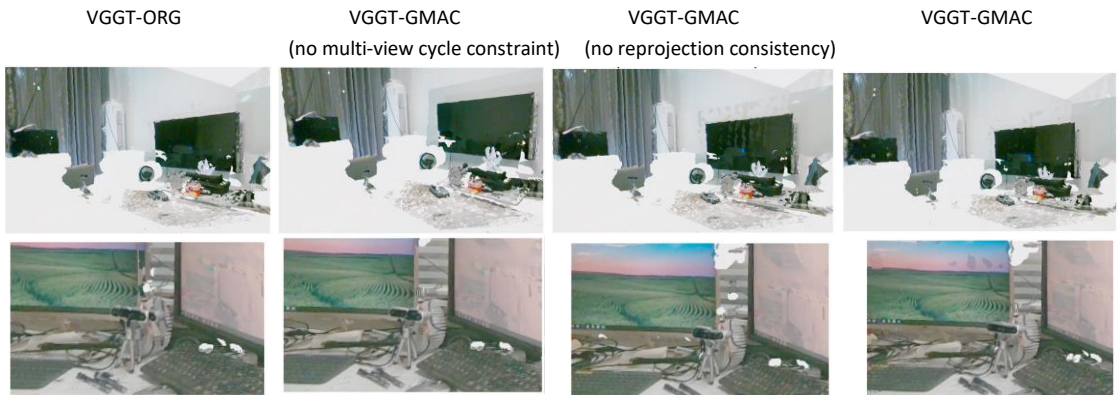

Fig. 1 The extrinsic parameter estimation performance of VGGT prediction versus the performance after sequentially removing relevant components of GMAC

Since the benchmark dataset provides accurate ground-truth extrinsics under ideal imaging conditions, these results reflect the upper-bound performance of the proposed method. They also indicate that the GMAC method can effectively exploit the implicit geometric information learned by multi-view networks, without relying on a specific backbone architecture.

b. Results on Real-World Data with Different Backbones

As reported in Table II, the proposed method consistently outperforms the baseline without geometric constraints on real-world data for both backbone networks. Both rotation and translation errors are substantially reduced, demonstrating that the method remains effective under challenging conditions such as complex illumination changes and dynamic scene interference.

Table II Results on the real-world multi-camera data.

| Backbone | Method | Rot. Error ↓ | Trans. Error ↓ |
| --- | --- | --- | --- |
| VGGSfM | Orginal | 2.89 | 6.9 |
| | Ours | 1.73 | 4.3 |
| MapAnything | Orginal | 2.75 | 6.6 |
| | Ours | 1.67 | 4.0 |
| VGGT | Orginal | 2.47 | 6.1 |
| | Ours | 1.51 | 3.6 |

c. Ablation Study

To analyze the contribution of each component to the overall performance, we conduct ablation studies on the real-world multi-camera dataset using VGGT as the backbone network. The experimental setup and evaluation metrics are kept identical to those used in the previous sections.

Specifically, we compare variants obtained by removing the reprojection consistency constraint (No RC) and the multi-view cycle constraint (No MC). The results are reported in Table III.

Table III Ablation results on the real-world dataset

| Method | Rot. Error ↓ | Trans. Error ↓ |
| --- | --- | --- |
| No RC | 2.11 | 5.2 |
| No MC | 2.01 | 4.8 |
| GMAC | 1.51 | 3.6 |

Removing the multi-view cycle consistency constraint leads to increased rotation and translation errors, indicating that the cycle consistency constraint captures higher-order geometric relationships among cameras and enhances the stability of global optimization. In contrast, removing the reprojection consistency constraint results in the most severe performance degradation, demonstrating that cross-view geometric consistency remains the key factor for accurate multi-camera extrinsic estimation.

We further test the visualization results after removing the two GMAC constraints. As shown in Fig. 1, removing the reprojection consistency constraint leads to noticeable local misalignment, removing the multi-view cycle consistency constraint causes global geometric distortion, while the complete method with both constraints preserved achieves nearly perfect cross-view point alignment and a stable geometric structure.

Overall, these results validate the complementary roles of the proposed components in improving both extrinsic accuracy: reprojection consistency ensures local precision, while the multi-view cycle consistency constraint enhances global consistency and optimization stability.

## IV. CONCLUSION

This paper presents a geometry-aware and backbone-agnostic framework for automatic extrinsic calibration in multi-camera systems. By reconfiguring and constraining pre-trained multi-view reconstruction networks, the proposed method enables accurate and stable camera extrinsic estimation without requiring explicit calibration targets or additional supervision.

Extensive experiments on both synthetic and real-world datasets, as well as across different backbone networks, demonstrate that the method does not rely on a specific network architecture and consistently improves extrinsic estimation accuracy, highlighting its high flexibility and modularity.

Overall, this work provides a practical and general solution for online or target-free multi-camera extrinsic calibration, and it can be readily integrated into existing multi-view perception and reconstruction systems.